\def\year{2021}\relax
\documentclass[letterpaper]{article} 
\usepackage{aaai21}  
\usepackage{times}  
\usepackage{helvet} 
\usepackage{courier}  
\usepackage[hyphens]{url}  
\usepackage{graphicx} 
\def\UrlFont{\rm}  
\usepackage{natbib}  
\usepackage{caption} 
\usepackage[utf8]{inputenc} 
\usepackage[T1]{fontenc}    
\usepackage{url}            
\usepackage{booktabs}       
\usepackage{amsfonts}       
\usepackage{nicefrac}       
\usepackage{microtype}      
\usepackage{amsmath}

\usepackage{graphicx}
\usepackage{amsmath,amssymb}
\usepackage{bm}
\usepackage[]{algorithm2e}
\usepackage{multirow}
\usepackage{breakurl}
\usepackage{color}
\usepackage{natbib}
\usepackage{soul}
\usepackage{wrapfig}

\title{Continual General Chunking Problem and SyncMap}
\author{
    Danilo Vasconcellos Vargas\textsuperscript{\rm 1,}\textsuperscript{\rm 2},
    Toshitake Asabuki\textsuperscript{\rm 3}
    \\
}
\affiliations{
    \textsuperscript{\rm 1}Kyushu University, Fukuoka, Japan\\
    \textsuperscript{\rm 2}The University of Tokyo, Tokyo, Japan\\
    \textsuperscript{\rm 3}Okinawa Institute of Science and Technology Graduate University, Okinawa, Japan\\
    vargas@inf.kyushu-u.ac.jp, toshitake.asabuki2@oist.jp

}

\begin{document}

\maketitle

\begin{abstract}
Humans possess an inherent ability to chunk sequences into their constituent parts. In fact, this ability is thought to bootstrap language skills and learning of image patterns which might be a key to a more animal-like type of intelligence. Here, we propose a continual generalization of the chunking problem (an unsupervised problem), encompassing fixed and probabilistic chunks, discovery of temporal and causal structures and their continual variations. Additionally, we propose an algorithm called SyncMap\footnote{Code available at https://github.com/zweifel/SyncMap} that can learn and adapt to changes in the problem by creating a dynamic map which preserves the correlation between variables. Results of SyncMap  suggest that the proposed algorithm learn near optimal solutions, despite the presence of many types of structures and their continual variation. When compared to Word2vec, PARSER and MRIL, SyncMap surpasses or ties with the best algorithm on $66\%$ of the scenarios while being the second best in the remaining $34\%$. SyncMap's model-free simple dynamics and the absence of loss functions reveal that, perhaps surprisingly, much can be done with self-organization alone.
\end{abstract}

\section{Introduction}
Humans are able to rapidly detect patterns in sequences \citep{Bekinschtein1672,StraussE1353}.
By detecting and chunking together patterns found, even without a supervised signal, humans are able to classify sounds, images and other information signals \citep{bulf2011visual}, \citep{orban2008bayesian}.
Therefore, this unsupervised learning process is thought to bootstrap many of the initial cognitive capabilities such as natural language processing, speech recognition and even image recognition.

Here, motivated by this general learning ability we propose the continual general chunking problem.
To evaluate the quality of an algorithm in the proposed problem a set of tests are defined, including chunks with fixed time series, chunks based on graphs with probabilistic transitions, overlapping chunks, chunks with probabilistic cycles, temporal and causal structures identification and continual scenarios. 
In other words, the continual general chunking problem generalizes the chunking problem to the identification of temporal and causal structures.
All this taking into consideration continual learning (change of the underlying structure as well as the probabilistic distribution of variables throughout the experiment).
Thus, the proposed problem joins the neuroscientific/psychologic chunking problem to the discovery of causal/temporal structure and unsupervised feature learning of time series.

To tackle the continual general chunking problem, we propose an algorithm, that without any parameter adjustment between problems, works for all the continual general chunking problems and achieves near optimal solutions. 
In other words, it can cope with probabilistic and fixed chunks as well as causal structures.
This algorithm is inspired by Hebbian learning and negative feedback signals.
It works by first converting inputs into spikes with slow decaying rate and creating a map on which signals self-organize through neuron group attraction/repeal forces. 
By creating a dynamic in which signals that activate together or deactivate together are attracted to a common center, the self-organizing dynamics are able to create clusters of correlated signals. 
It is worth noticing that attraction of the activated signals (which is closely related to Hebbian learning) by itself is not enough.
Attraction of both activated and deactivated signals are necessary for the dynamics to reach the cohesive behavior described here.

\section{Our Contributions}
In this paper, a general problem is proposed called Continual General Chunking Problem as well as an algorithm to solve it called SyncMap.
The key contributions are as follows:
\begin{itemize}
	\item We generalize various problems from neuroscience and computer science (i.e., chunking problem, discovery of causal/temporal communities, unsupervised feature learning of time series and their continual variations) into a problem called Continual General Chunking Problem (CGCP).
	CGCP is defined formally and experiments are developed to evaluate an algorithm's performance.
	Chunking problem alone encompasses problems from learning image features to sounds as shown in \cite{asabuki2020somatodendritic}. 
	It originates from detecting repetitive patterns of neural spike sequences, but its primitives are thought to be widely used in the brain. 
	Discovery of causal and/or temporal communities was explored in \cite{schapiro2013neural} with applications in \cite{jiao2018exploring}.
	Here we shown how all these problems and their continual variations can be seen as a single one.
	\item We propose an algorithm for tackling CGCP called SyncMap. 
	SyncMap is a different type of self-organizing map with dynamics of attraction between all nodes that activate or deactivate at the same time. 
	It shares with other self-organizing systems such as Self-Organizing Map \cite{kohonen2013essentials} and Novelty Map \cite{vargas2015novelty} only the idea of using a map as all other dynamics and intent differ.
	Moreover, its self-organizing dynamics enables it to flexibly respond to changing environments which is a challenge for most algorithms that optimize loss functions.
	\item Beyond generalizing the problems, we consider (perhaps for the first time) continual variations of them in the CGCP.
	The challenge here is to respond quickly to the environment, adapting previous learned structures and correlations that have changed.
	This is motivated by the constant adaptation spotted in neural cells which can rapidly switch behavior according to environmental changes \cite{dahmen2010adaptation}.
	\item Experiments on fixed chunks, probabilistic chunks and temporal structures suggest that SyncMap reaches near optimal solutions.
	The same is true for continual variations of them, i.e., when such probabilistic chunks or temporal structures change throughout the experiment.
	Moreover, we extend the tests for detecting temporal structures of real world scenarios.
\end{itemize}

\section{Related Work}

\subparagraph{Chunking.} Through the process, called “chunking”,  the brain attains compact representation of sequences, which is thought to reduce the complexity of temporal information processing and associated cost \citep{ramkumar2016chunking}.  Chunking in the brain computation is crucial to achieve high-order functions that require hierarchical sequence processing, such as motor-skill learning \citep{graybiel1998basal,jin2014basal} and language acquisition  \citep{buiatti2009investigating,gentner2006recursive}. Cognitive experiments suggests that children learn words as chunks \citep{Saffran2003}. This process is thought to contribute to higher-order learning process, and be fundamental mechanisms that children identify words from speech \citep{dehaene2015neural,hay2011linking}.  Recent study with human subjects has shown that chunking occurs even when the sequence has co-occurring structure rather than a repetitive pattern \citep{schapiro2013neural}. The influential chunking (word segmentation) algorithm, PARSER has been proposed, which extract the all frequently appeared n-grams within the sequence \citep{perruchet1998parser}. Despite PARSER works well for simple chunking tasks and computational linguistic applications \citep{goldwater2009bayesian, feldman2013role}
, yet fail to detect chunks if the transition probability over all elements are uniform\citep{schapiro2013neural}. Recently, biology-inspired sequence learning model, called Minimization of Regularised Information Loss (MRIL) has been proposed \cite{asabuki2020somatodendritic}. MRIL is applicable to broad range of chunking task including uniform transitions. These studies suggest that chunking is a fundamental process, yet the mechanism of which is still elusive. Albeit the multitude of applications of chunking and the wide interest of neuroscientists on the subject, this subject was not fully explored from a machine learning perspective. 
Chunking can be found in some articles focusing on natural language processing.
Sometimes chunking is modeled as a supervised pre-processing step \citep{zhai2017neural}. 
In other cases, it is used for unsupervised grammar induction \citep{ponvert2011simple}.
In both cases the problem taking into account differs from the original task independent one defined in neuroscience. 

\subparagraph{Unsupervised Learning for Sequences.} Unsupervised learning for sequences usually extract features which can predict future input \citep{clark2019unsupervised,hyvarinen2016unsupervised,lei2019similarity,mikolov2013efficient,wu2018hierarchical}.
These features, albeit descriptive of the sequences and useful for classification of sequences, do not uncover the chunks present.
Related to unsupervised learning for sequences, contrastive predictive encoding (CPE) is a peculiar learning algorithm which makes use of a probabilistic contrastive loss, inducing the latent space to encode maximally useful information in its representation \citep{hjelm2018learning}.
It requires the sequence of samples to have some sort of order and could use a general chunking algorithm to present images that are coherent to a certain class.
In this manner, CPE and its applied problem formulation is complementary rather than competing with the problem proposed here.
Moreover, self-organizing maps and their variations tackle a related but different problem \citep{fortuin2018som}. 
They can learn topological representations of time series and static data while chunks are temporal correlations between variables, therefore, their intent differ.

\subparagraph{Latent Variable Estimation.}
Latent variable estimation \citep{fox2011sticky, pfau2010probabilistic, qian2014learning, rabin1963probabilistic} resemble some of the chunking problems defined here. However, their objective differ. Chunking is a clustering over the variables of the problem respecting their temporal correlation. Even if chunks of variables are abstracted as meta-variables (a set of variables) there is still an inherent difference between chunks and latent variables. 

\subparagraph{Word Embeddings.} To transform words and paragraphs into vectors of numbers, word embeddings are used in natural language processing \cite{mikolov2013efficient}, \cite{khattak2019survey}. 
Some of them are enriched with information specific to natural language processing such as FastText \cite{fasttext2017} or are contextualized word embeddings such as ELMo \cite{peters2018deep} and BERT \cite{devlin2018bert}. However, prediction-based word2vec embeddings \cite{mikolov2013efficient} and co-occurrence matrix based GloVe embeddings \cite{pennington2014glove} can be also used for more general problems similar to the one presented in this paper.

\subparagraph{Causal Graphs.} As part of the general chunk problem, this paper aims to discover temporal and causal structures in sequences which are related to causal graphs.
The detection of temporal structures has many practical applications and can be used as well for facilitating the learning of causal graphs by identifying confounders, identifying correlated variables, among others \citep{cai2019triad,kocaoglu2019characterization}.
Interestingly, some terms from causality studies also take place in chunking problems albeit in different ways.
For example if variables $x$ and $y$ are confounded by a given variable $z$, if both $x$ and $y$ pertains to a different chunk, $z$ is an overlapping variable.
This overlapping variable can be identified as a different chunk which should facilitate the learning of causal graphs.

\section{Continual General Chunking Problem}

We define continual general chunking as the problem of extracting co-occurring states from a time series of which the underlying generation process changes depending on the task as well as throughout the task. 
Here, we first describe the input time series structure considered in this problem.

Our input sequence consists of state variables, transitioning by first-order Markov chain, of which the element of transition matrix is defined as: $P_{ab}=Pr[s_{t+1}=b|s_{t}=a]$, where $s_t$ is the state variable at time $t$; and $a$ and $b$ are the labels of states. Here, note that each state $s_t$ is a vector. In our sequence, each state belongs to either a fixed chunk or a probabilistic one. In a fixed chunk, given the state at the current time, the state at the next time step is selected deterministically. Let $a$ and $b$ be elements of a fixed chunk, with the direction $a$ to $b$, then the transition matrix should satisfy: $P_{ab}=1$. Note that since $\Sigma_{b} P_{ab}=1$, the state $i$ has only one possible next state.

In real situations, the deterministic chunks explained above is unrealistic, and the stochastic transition even within the chunks should be allowed. In order to include such cases, let us first define a structure of the input sequences. Our input can be generated by a random walk over graphs characterized by the two types of degrees- internal and external degrees. The internal degree $k_a^{int}$ of state $a$ is the number of connections with nodes that belong to the same chunk. The external degree $k_a^{ext}$ is the number of links that node $a$ connects to nodes belonging to other chunks. For all states, we assumed that the following relation holds:

\begin{equation}
k_a^{int}>k_a^{ext} (\forall a \in A),
\end{equation}
where $A$ is the set of states in the sequences. The above relations are satisfied if the graph has dense connections within chunks yet sparse connections between nodes in different chunks. The graph structure defined above is also known as communities \cite{Radicchi2658}. In our setting, the transition probabilities were assigned to all links, hence stochastic transitions were allowed in all connected states. Note that both $P_{ab}$ and $P_{ba}$ can be assigned in a link between a pair of states. A special case of our case was investigated in a cognitive experiment \citep{schapiro2013neural}. In their setting, they assumed that transition probabilities $P_{ab}$ are uniform over all states, which is not the case in our setting.

In our continual setting, the causal structure is behind the generation process, hence the transition matrix can change over time. In this paper, we assume that the input sequence can take $m$ graph structures. The label of transition matrixes, which determines the  causal structure, is termed as "tasks". Assume the set of transition matrixes, ${\bf P}=\{P^{(1)}, P^{(2)},\ldots, P^{(m)}\}$. At the time when the tasks are switched, one task is randomly selected from the set ${\bf P}$, and the time series is continuously generated. We set $m=2$ and since the generation process itself changes, the set of state variables that constitute each chunk also change.

\section{SyncMap}

Inspired by how Hebbian volume learning complements Hebbian, taking into consideration the effects of nearby neurons learning \cite{mitchison1999can} as well as how feedback systems affect the learning of algorithms for chunking problems \cite{asabuki2020somatodendritic}; here we propose a map in which neurons in a group can learn together to represent interrelated concepts .
The main idea here is to use a simplified Hebbian learning together with feedback dynamics to create a projection that encodes the probability of two variables activating at the same time with their distances in this new generated space, i.e., encoding the relative probabilistic correlation between variables as distances between them.
Thus, variables that activate together will have their respective projections drawn to their middle point while variables that do not activate together will be, consequently, pulled away from each other.

SyncMap is divided into two steps: (a) the activation of nodes and their update inside the map (b) a clustering phase on the learned map, revealing their clusters.
The following subsections explain these steps in detail.

\subsection{Input Encoding}
\label{input_method}

Let $s_{t} = {s_{1,t}, s_{2,t}, ..., s_{n,t}}$ be the set of state values at time $t$ for $s_t \in \{0,1\}^n:\sum^n_{i=1} s_{i,t} = 1$. 
The input encoding is an exponentially decaying vector $x_t$ with the same size as the number of states:
\begin{equation}
x_{i,t} = 
\begin{cases}
    s_{i,ta}*e^{-0.1*(t-ta)},     	&ta < m*tstep\\
    0,      &otherwise
\end{cases}
\end{equation}
where $ta$ is the most recent state transition to state $s_i$.
State transitions happen every $tstep$ steps and variables which have their time of activation greater than $m*tstep$ are set to $0$.
Consequently, the system can only remember the last $m$ variables presented.
All experiments here are conducted using $m=2$ and $tstep=10$.

This representation of input can also be encoded differently without any change in result, e.g., by storing the last $m$ inputs directly. 
In other words, the details of implementation for the input is not important for the method itself.
It suffices to remember the last $m$ states.

\subsection{Dynamics}
\label{sec_dynamics}

First all inputs $x_{i,t}$ have a corresponding weight $w_{i,t}$ initialized to a random position in the map $w_{i,t} \in \mathbb{R}^k$, creating a pair $(x_{i,t}, w_{i,t})$.
$w_{i,t}$ is a $k$ dimensional variable and $k$ is a parameter defined a priori (the dimension of the map). 

Every time a new input $x_t$ is presented, each of its constituents $x_{i,t}$ are divided into activated or recently activated (positive) and non-recently activated (negative) inputs.
Specifically, all inputs are converted into two sets: positive $PS_t$ and negative $NS_t$ sets.
Inputs with value greater than $0.1$ are a member of $PS_t$.
Otherwise, inputs are a member of $NS_t$.
In other words, the following holds true:
\begin{eqnarray}
 PS_t = \{ i  | x_{i,t} > 0.1\}\\
 NS_t = \{ i  | x_{i,t} \leq 0.1\}
\end{eqnarray}

If the cardinality of both sets are greater than one, i.e. $|PS_t| > 1$ and $|NS_t| > 1$; then the center of both sets are calculated as follows:
\begin{eqnarray}
cp_t = \frac{\sum_{i \in PS_t} w_{i,t}}{|PS|} \\
cn_t = \frac{\sum_{i \in NS_t} w_{i,t}}{|NS|}, 
\end{eqnarray}
in which $cn_t$ and $cp_t$ are respectively the centers of $NS_t$ and $PS_t$ sets.
If the cardinality of both sets are not greater than nothing is updated in this iteration.
After the center of both sets are calculated, the position $w_{i,t}$ of each input is updated.
\begin{eqnarray}
\phi(i,t) = 
\begin{cases}
    1,     	&i \in PS_t\\
    0,      &i \in NS_t
\end{cases}\\
w_{i,t+1} = w_{i,t} + \alpha*(\frac{\phi(i,t)*(cp_t-w_{i,t})}{\left\| w_{i,t} - cp_t \right\|} -\\   \frac{(1-\phi(i,t))*(cn_t-w_{i,t})}{\left\| w_{i,t} - cn_t \right\|})
\end{eqnarray}
where $\alpha$ is the learning rate.
Subsequently, all values of $w_{i,t+1}$ are normalized to be within a hypersphere of radius $r=1$.

\subsection{Clustering}

The clustering step happens after the dynamics described in the previous section is repeated for each input. 
In the clustering step, the projected map $w$ is clustered to determine effectively the chunks/communities.
Here, DBSCAN \cite{schubert2017dbscan} is used for this procedure because it does not require the number of clusters as input.
The required parameter is the density of clusters $eps$, which is somewhat fixed for a given hypersphere of radius $r$, and the minimum size of each cluster $mc$ which can also be set independent of problem.
$eps$ and $mc$ are set to respectively $3$ and $2$.
Having said that, other clustering algorithms together with the use of clustering analysis techniques for discovering number of clusters should be equally or even more effective. 
For simplicity we are narrowing the scope of this paper to DBSCAN alone.

\section{Experiments}

The experiments compose a total of nine different tests encompassing fixed chunks, mixed structures, their continual variations, long chunks, overlapped chunks and real world scenarios. Both overlapped chunks and real world scenarios have two distinctive tasks.

\subparagraph{Settings.}
In all experiments, the learning algorithm is first exposed to $100000$ samples of the problem and followed by an extraction of the chunks predicted. 
This is also true for continual variations, where the problem changes after $100000$ samples are inputted, with the second problem also presenting $100000$ samples before the run is finished.
All tests are run on a MacBook Pro 10.15.5 2.3Ghz 16GB laptop as they demand little computational effort. 

Here we compare four distinct algorithms: SyncMap (proposed one), MRIL \cite{asabuki2020somatodendritic}, PARSER and Word2vec. 
SyncMap's parameters $\alpha$ and $k$ are fixed to respectively $0.1$ and $3$.
Regarding the PARSER algorithm, since it finds possible n-grams, hence not whole chunks, we first excluded the unnecessary long n-grams ($n>6$), and concatenated the rest of the short segments, of which share the same element. These resultant segments were regarded as "chunks" that PARSER extracted.
To evaluate how a word embedding algorithm would fair in CGCP, we include a skip-gram Word2vec embedding modified to work in the context of CGCP.
A dense deep neural network model was used as model for the Word2vec with a latent dimension of $3$ and an output layer with softmax and size equal to the number of inputs. The chosen training parameters are $10$ epochs, $1e-3$ learning rate and $64$ batch size with a mean squared error as loss. These parameters were the best performing without unnecessary slowdown after a dozen trials. A window of $100$ steps was used to calculate the output probability of skip gram. The input was kept the same as SyncMap, since variations of non-decaying ones performed (perhaps surprisingly) worse. Therefore, the window for Word2vec include $10$ state transitions; equivalent of $100$ time steps. 
Regarding MRIL, we used five output neurons for all tasks, with the learning rate $1e-3$. For comparison, we used the same decaying input as SyncMap, rather than the Poisson spiking input used in the original setting of MRIL. We grouped the output neurons showing correlation larger than $0.5$, and determining chunk by assigning an index of groups that maximally respond to each input.

\begin{figure*}[h]
\centering
 \includegraphics[width=0.12\textwidth]{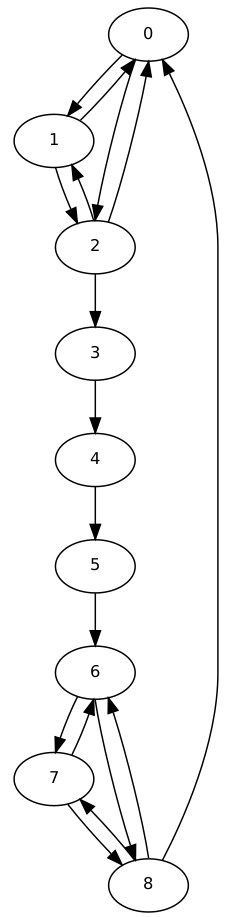}
 \includegraphics[width=0.4\textwidth]{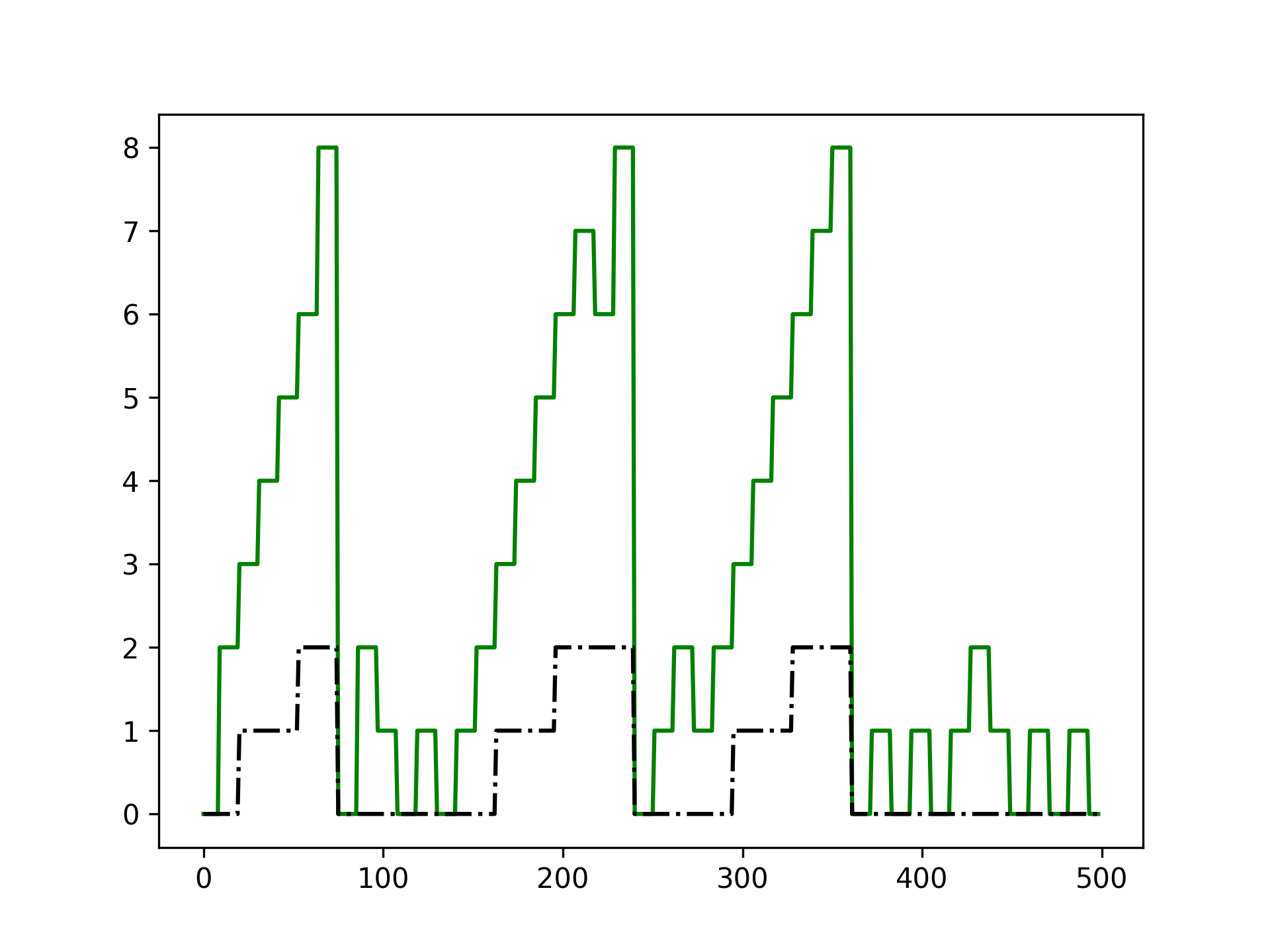}
 \includegraphics[width=0.13\textwidth]{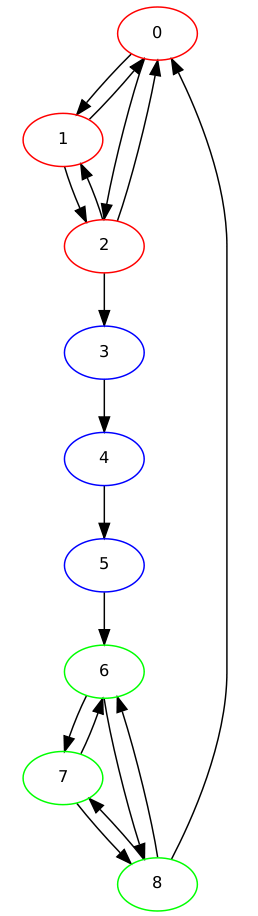}\\
 \caption{
Problem description (up left), learned input-output map (upper middle) and respective learned clustering (upper right).
The problem is a cyclic directed graph with node numbers as indicated.
Each node number consists of a given input which is activated and inputted by a random walk over the graph (see the Section~\ref{input_method} for more details).
The learned input-output map is as follows: input (red line) and learned output (dash-dotted line).
In direct correspondence to the learned output, it is possible to cluster the nodes with the colors shown (upper right). The learned map is shown in Fig~\ref{mixed_problem2}.
}
 \label{mixed_problem}
\end{figure*}

\subparagraph{Fixed Chunks.}
The problem considered here has four fixed chunks each containing three different variables. 
Transition between chunks happen at the end of the chunk sequence, i.e., after the third variable inside the chunk is presented. 
A chunk can transition to any other chunk with equal probability.

\subparagraph{Overlapped and Long Chunks.}
In one hand, overlapped chunks evaluate the capability of systems to perceive chunks that share some variables. 
On the other hand, long chunks evaluate if the frequency of chunks affect the system.
Both overlapped and long chunks are probabilistic chunks.

For the overlapped chunks, two problems are tested: overlap1 and overlap2.
Overlap1 have two chunks composed respectively of variables a,b,c,d,e and d,e,f,g,h while overlap2 has chunks with respectively variables a,b,c,d,e and a,b,c,d,e,f,g,h,i,j. In other words, they either share variables of are a subset of each other.
For the long chunk experiment, each chunk has four non-repeating variables a,b,c,d and e,f,g,h presented randomly each step, however, the duration of the first chunk is four transitions while the second has a stochastic duration of $5 + unif(0,20)$ transitions. $unif(min,max)$ defines a uniform distribution within the half-open interval $[min,max)$.

\subparagraph{Mixed Structures.}
In this problem, both probabilistic chunks together with fixed one are present in the system.
The graph with the structure and transitions of the problem is shown in the left of Fig.~\ref{mixed_problem}. It has 2 probabilistic chunks and one fixed chunk.

\begin{figure}[h]
 \centering
 \includegraphics[width=0.85\linewidth]{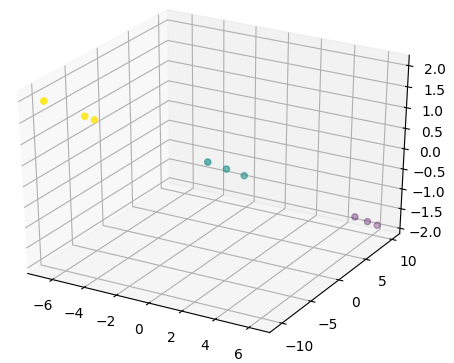}
 \caption{The learned map by SyncMap together with colors showing the chunks learned for problem described in Fig\ref{mixed_problem}.}
 \label{mixed_problem2}
\end{figure}

\begin{figure*}[h]
\centering
\includegraphics[width=\linewidth]{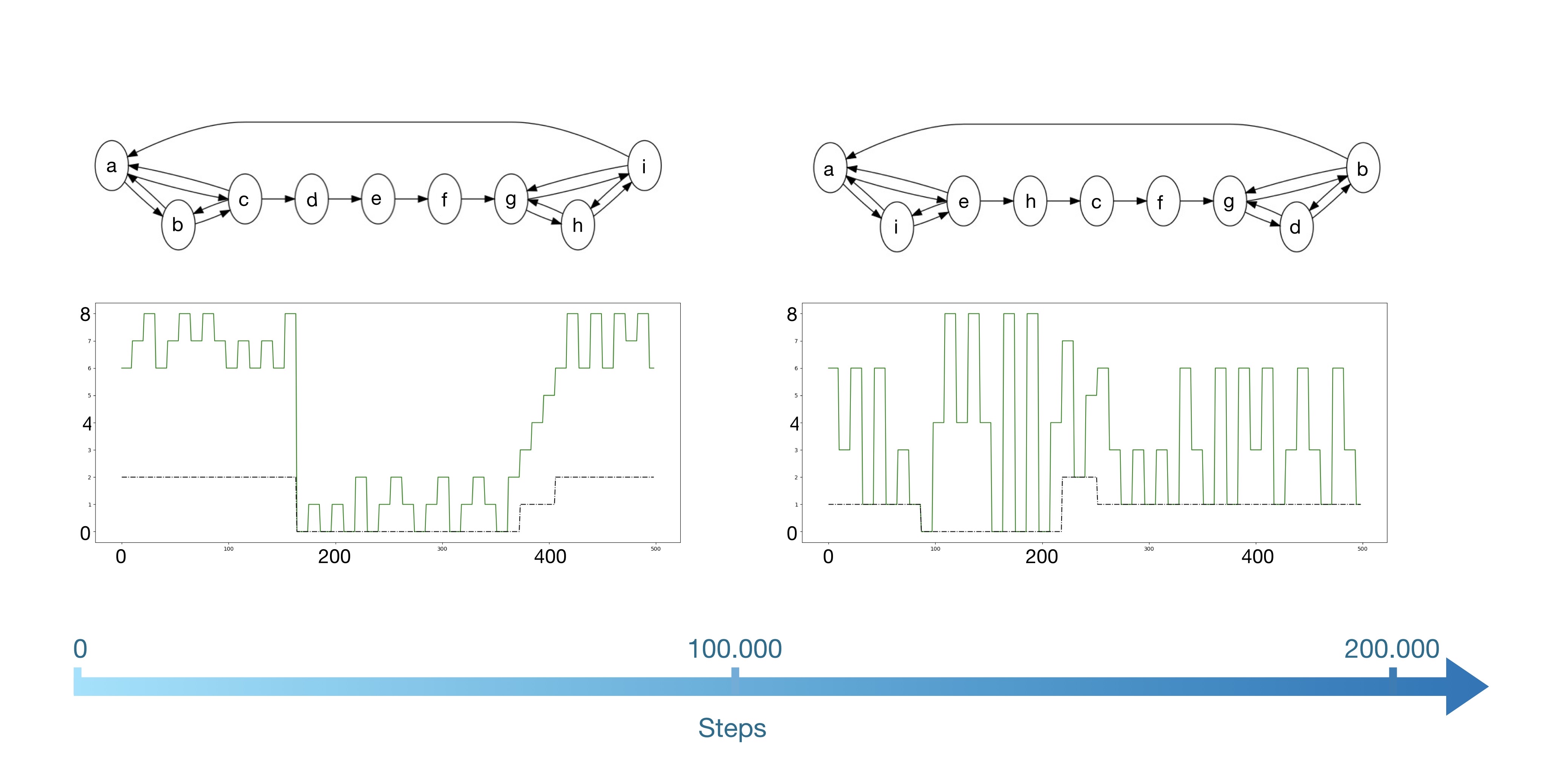}
 \caption{
Illustration of the continual version of the mixed structures problem and an example of SyncMap's learning output. The structure changes after $100000$ steps and then the experiment ends after $200000$ steps. The upper graphs show the start (left) and end (right) problem structures and their respective variables. The start and end outputs of SyncMap are shown at the bottom.  
}
 \label{continual_mixed}
\end{figure*}

\subparagraph{Continual Variations.}
Two continual variations of previous problems are proposed: continual fixed and continual mixed.
They are variations of respectively the fixed chunk and the mixed structures. In both variations, variables are permuted between chunks when tasks are changed. 
For the continual variation of the fixed chunk, in the first task the configuration of chunks is (a,b,c), (d,e,f), (g,h,i), (j,k,l). In the second task, the fixed chunks become respectively (a,k,i), (g,e,j), (d,h,c) and (f,b,l).
Regarding the continual mixed problem, Fig.~\ref{continual_mixed} shows how the structure of the problem changes throughout.

\subparagraph{Real World Scenarios.} We test in two variations of a real world scenario. 
Specifically, the recognition of probabilistic chunks in the first-order Markov model of theme transitions for humpback whales' song types \cite{garland2017song}. 
Since the transition between nodes are not given, they are considered equally probable. 
Sequence 1 and 2 are defined as respectively the graphs A and B from Fig~1 in the supplementary materials.

\section{Results and Analysis}

\begin{table*}[h]

  \centering
  \begin{tabular}{lccccc}
    \toprule
    {\bf Model}     & {\bf Fixed Chunks}       & {\bf Long Chunks}  &{\bf Mixed Structures}\\
    \midrule
    PARSER & 0.95$\pm$0.07  & 0.16$\pm$0.27  & 0.36$\pm$0.40 \\
    \midrule
    Word2vec & {\bf1.0$\pm$0.0}  & 0.68$\pm$0.22  & 0.78$\pm$0.07 \\
    \midrule
    MRIL & 0.86$\pm$0.13  & 0.76$\pm$0.17  & 0.85$\pm$0.16 \\
    \midrule
    {\bf Ours}: SyncMap & 0.97$\pm$0.09 & {\bf0.97$\pm$0.06} & {\bf0.95$\pm$0.06}\\
    \bottomrule
  \end{tabular}
    \caption{Mutual information comparison over SyncMap, PARSER, Word2vec and MRIL in Fixed chunks, Long chunks and Mixed structures settings. Best mean and statistically similar results (results with p<0.05 in a t-test with null hypothesis of having the same mean as the best mean) are in bold.}
  \label{fixed_long_mixed}
\end{table*}
In this paper, we define the optimality of solutions by the degree of correlation with the ground truth. For all tests, as a correlation metric, we measured the normalized mutual information scores for Wor2vec, MRIL, PARSER and SyncMap (Tables~\ref{fixed_long_mixed}, \ref{continual_table} and \ref{overlap_sequence}). Here, the normalized mutual information score is defined as: 
\begin{eqnarray}
NMI(\hat{Y},Y) = 2\frac{I(\hat{Y};Y)}{H(\hat{Y})+H(Y)},
\end{eqnarray}
where $\hat{Y}$ and $Y$ are the output of algorithms and the ground truth, respectively. $I(\hat{Y};Y)$ is the mutual information between $\hat{Y}$ and $Y$, and $H(\cdot)$ is the entropy. This NMI measure takes value between a minimum of 0 (no mutual information) and a maximum of 1 (perfect correlation).

The proposed algorithm SyncMap performed better overall. It surpassed or performed within the standard deviation of other algorithms in $7$ out of the $9$ tests. PARSER and Word2vec performed similarly, but had only $3$ out of the $9$ results that were comparable with the others. MRIL was better in only one of the tests but if SyncMap is removed from the comparison it performs slightly better than both PARSER and Word2vec.
In other words, SyncMap and MRIL are both general algorithms while Word2vec and PARSER have some specific problems they are very good at. 
\begin{table*}[h]
  
  \centering
  \begin{tabular}{lccccc}
    \toprule
    \multirow{2}{*}{\bf Model} &
      \multicolumn{2}{c}{\bf Continual fixed} &
      \multicolumn{2}{c}{\bf Continual mixed}\\ 
      \cline{2-5}
    & {Task 1} & {Task 2} &{Task 1} &{Task 2}\\
    \midrule
    PARSER &   0.97$\pm$0.06  &  {\bf0.96$\pm$0.07} &  0.47$\pm$0.40 & 0.28$\pm$0.38 \\
    \midrule
    Word2vec &   1.0$\pm$0.0 & 0.29$\pm$0.20&  0.76$\pm$0.08 & 0.0$\pm$0.0 \\
    \midrule
    MRIL &  0.80$\pm$0.16 & 0.53$\pm$0.12&  0.85$\pm$0.15 & 0.32$\pm$0.16 \\
    \midrule
    {\bf Ours}: SyncMap  & 0.93$\pm$0.13 &  0.89$\pm$0.15 & 0.97$\pm$0.04 & {\bf 0.99$\pm$0.03}\\
    \bottomrule
  \end{tabular}
  \caption{Mutual information comparison over SyncMap, PARSER, Word2vec and MRIL in continual fixed and continual mixed settings. In task 2, Best mean and statistically similar results (results with p<0.05 in a t-test with null hypothesis of having the same mean as the best mean) are in bold.}
  \label{continual_table}
\end{table*}

\begin{table*}[h]

  \centering
  \begin{tabular}{lcccc}
    \toprule
    {\bf Model}     & {\bf Overlap1} & {\bf Overlap2} &{\bf Sequence1} &{\bf Sequence2}\\
    \midrule
    PARSER &  {\bf 0.77$\pm$0.42} & 0.30$\pm$0.46 & 0.27$\pm$0.11 & 0.57$\pm$0.1\\
    \midrule
    Word2vec  & 0.21$\pm$0.28 & 0.06$\pm$0.09 &  0.28$\pm$0.03 & {\bf 0.79$\pm$0.08}\\
    \midrule
    MRIL  & 0.03$\pm$0.18 & 0.0$\pm$0.0 & {\bf  0.38$\pm$0.11} &0.51$\pm$0.10\\
    \midrule
    {\bf Ours}: SyncMap  & {\bf 0.70$\pm$0.19} & {\bf 0.64$\pm$0.10} &  {\bf 0.39$\pm$0.16} &0.61$\pm$0.06\\
    \bottomrule
  \end{tabular}
    \caption{Mutual information comparison over SyncMap, PARSER, Word2vec and MRIL in Overlapped chunks and Real world scenarios. Sequence1 and sequence2 correspond to parts A and B in Figure 1 in the supplementary materials, respectively (problem defined in \protect\cite{garland2017song}). Best mean and statistically similar results (results with p<0.05 in a t-test with null hypothesis of having the same mean as the best mean) are in bold.}
  \label{overlap_sequence}
\end{table*}

Regarding SyncMap, it performs similarly for long chunks and mixed structures (Table~\ref{fixed_long_mixed}).
Long chunks frequency is less of an issue because once variables are sufficient far away in the projected map, the update becomes weaker as well as deactivating together has also a similar attraction force.
Moreover, SyncMap considers only one-to-one correlations and create groups through the emergent attraction/repulsion behavior, consequently, the nature of the chunk or structure do not matter in this regard. 
Continual variations of the problems (Table~\ref{continual_table}) suggests that SyncMap is capable of adapting to changes in the structure without any observed deleterious effect.
This is expected since SyncMap steadily updates the correlation between variables projected into the map $w$. Once the dynamics reach an equilibrium the updates affect less the map distribution, however, a change in the underlying problem structure affects the place of attractors and therefore naturally put the system in an unstable state initiating the adaptation. 
SyncMap performs better than the other algorithms in overlapped chunks problems (Table~\ref{overlap_sequence}). However, there is still ground for improvement here as it cannot represent a hierarchical structure. 
In real world scenarios, SyncMap performs equally to all other in Sequence1 and is the second best in Sequence2. Increasing past state memory $m$ should enable better performance. 

Word2vec usually generates a map in which variables are more dispersed than the one produced by SyncMap which makes clustering difficult.
For long chunks and mixed structures, the map becomes fuzzier and therefore the MI score drops accordingly.
Moreover, it does not have a built-in adaptation which makes changes in the problem structure cause probabilities of nearby nodes to even out.
Big overlaps do a similar effect, probabilities of nearby nodes are similar, it ends up recognizing all nodes as a single chunk.
Real world sequences have mixed results, being the best in one and second worse on the other.


Regarding PARSER, it extracts chunks based on the bias of transition probabilities, therefore, performance deteriorates in time series with equal probability such as our long chunk and mixed structures.
In the continual variations, PARSER performs well in both tasks involving the fixed problem since forgetting phase during learning enables it to adapt to the new environment (i.e., second task).  Unlike the fixed case, it failed to learn in the mixed structures even in the first task, as shown previously.  
MI score for overlap2 becomes less than half of overlap1. We speculate this is because sequence of overlap2 has higher fraction of overlaps than sequence1. Therefore, the transition becomes much uniform which deteriorates the performance of PARSER. Since real world tasks have uniform transition part, the MI was lower than the fixed case.

Since MRIL detects spatio-temporal correlation over inputs, it showed better performance than PARSER and Word2vec if the sequence has uniform probability. MRIL learns not only the feedforward weights, but also lateral inhibition weight with the spike train correlation of output. Since each state in our sequence was presented only $10$ time steps, MRIL failed to detect spike correlation by its slower timescales, hence showed MI less than 0.9 in all problems. Similarly, the MI of the second task in continual problems was lower than that of first one because lateral inhibition could not be trained efficiently. In the overlap tasks, since MRIL creates chunks including both non-overlapped and overlapped states, the MI was significantly low. In the real world problems, the MI in sequence1 was almost comparable to that of SyncMap, yet lowest in sequence2.


\section{Conclusions}

In this paper, we proposed both a general problem called CGCP and an algorithm called SyncMap which outperforms or ties with competitive algorithms from neuroscience and machine learning in $7$ out of $9$  tests. 
We expect that variations of SyncMap should further better its performance relative to other algorithms in CGCP and will probably become a strong alternative in applications from natural language processing to image recognition.
Future directions will investigate problems with noise, hierarchy and causal relations as well as tasks specific to language processing and image/action recognition.

\section*{Acknowledgements}
This work was supported by JST, ACT-I Grant Number JP-50243 and JSPS KAKENHI Grant Number JP20241216.

\section*{Broader Impact}

In this paper, we proposed the continual general chunking problem which merges many problems from neuroscience and machine learning into a continual general one.
This should promote the exchange of knowledge between distant fields for both the development of new more animal-like intelligent algorithms as well as to aid in the understanding of intelligence.
The proposed algorithm, SyncMap, is an example of this with features inspired by neuroscience and the simplicity of machine learning.
Therefore, we believe that many researchers beyond machine learning may benefit from this research. 
There is nobody put at disadvantage with this research. 
Both the problem and the proposed solution are new developments which aim at increasing overall generality and robustness of AI systems. Failure is expected to decrease if more importance is giving to such general problems as proposed here.
CGCP aims at evaluating algorithms in many variations of problem types and settings. Therefore algorithms are not able to take advantage from biases even from specific problem classes.

\bibliography{general_chunking}

\end{document}


\maketitle

\section{Appendix A}

\begin{figure}[!htb]
\centering
 \includegraphics[width=\textwidth]{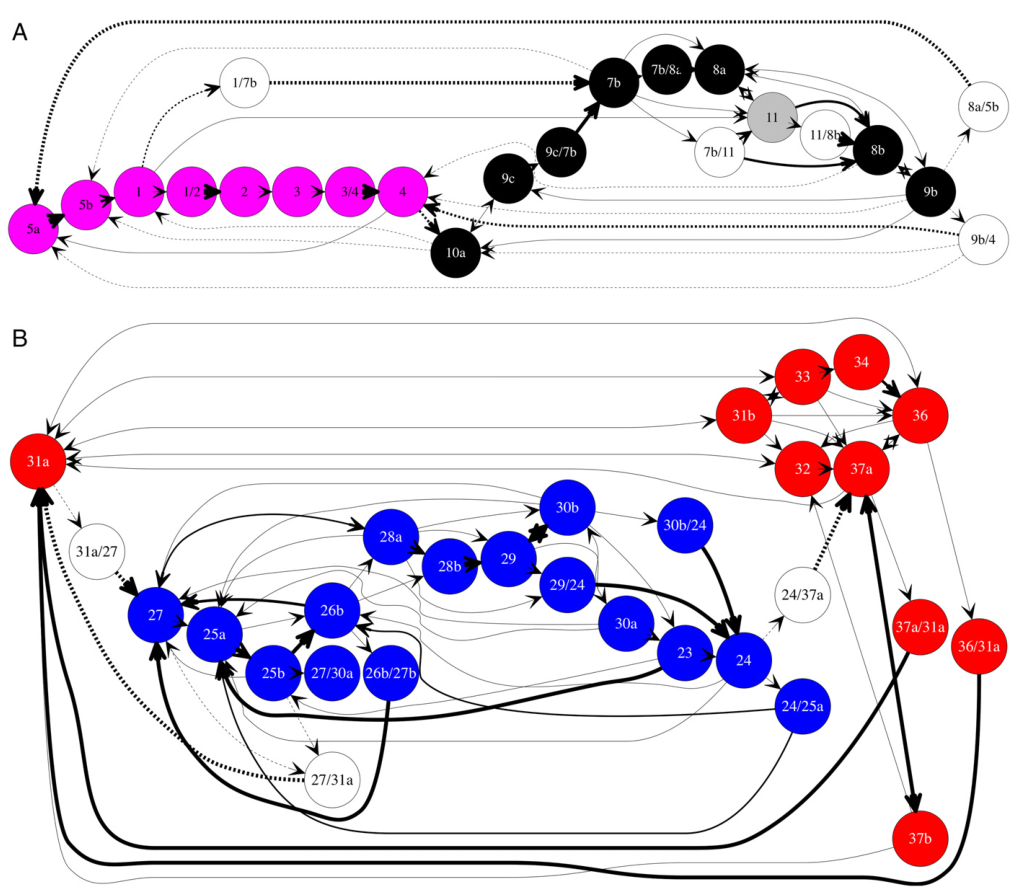}
 \caption{
First-order Markov Chain of humpback whales' song types \protect\cite{garland2017song}.
}
 \label{whalesong}
\end{figure}

Experiments on real world problems used the graph defined by Figure~\ref{whalesong}.

\section{Appendix B - Mapping Comparison (Encoding / Word Embedding)}

\begin{figure}[!htb]
\centering
 \includegraphics[width=0.48\textwidth]{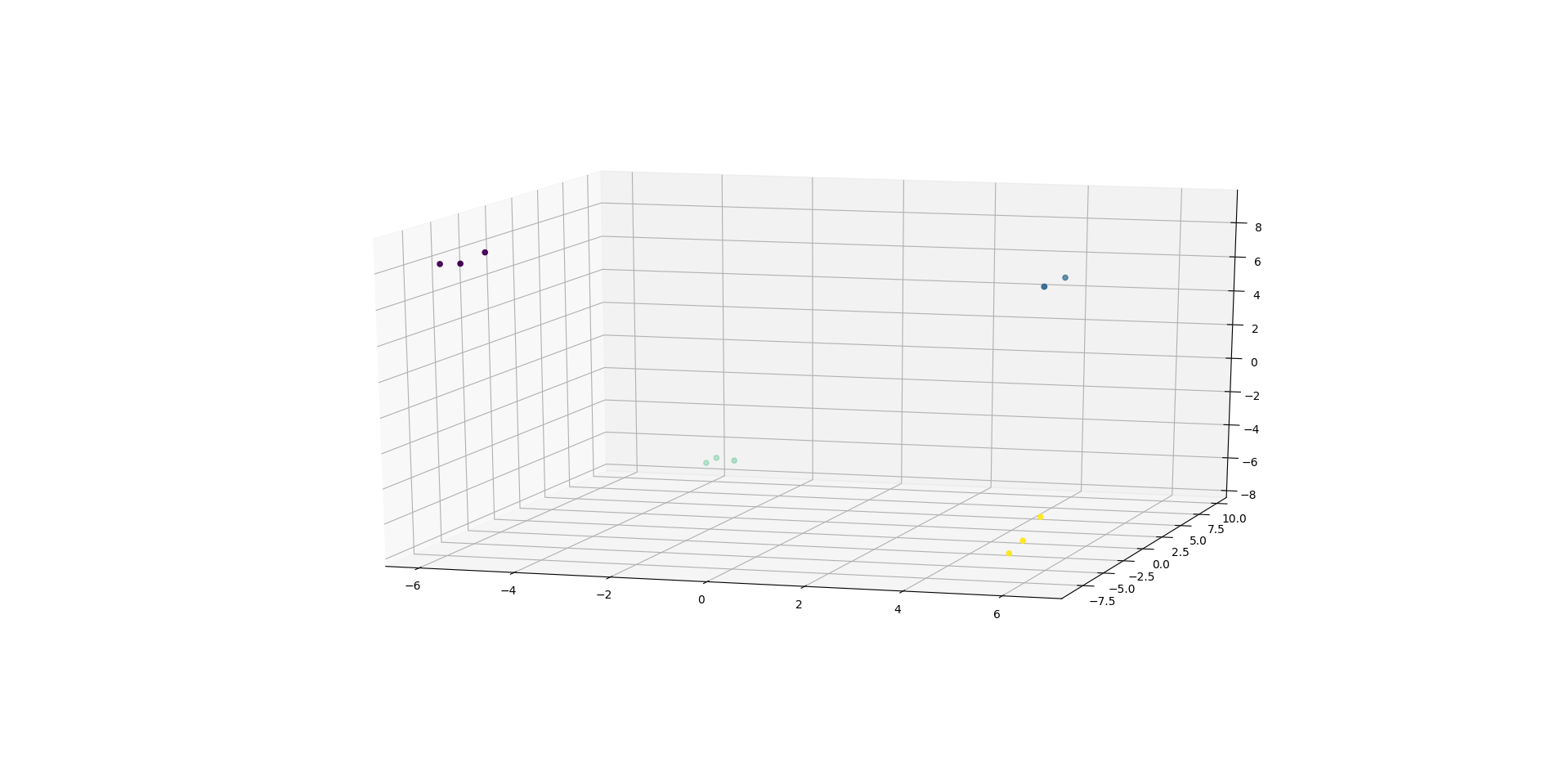}
  \includegraphics[width=0.48\textwidth]{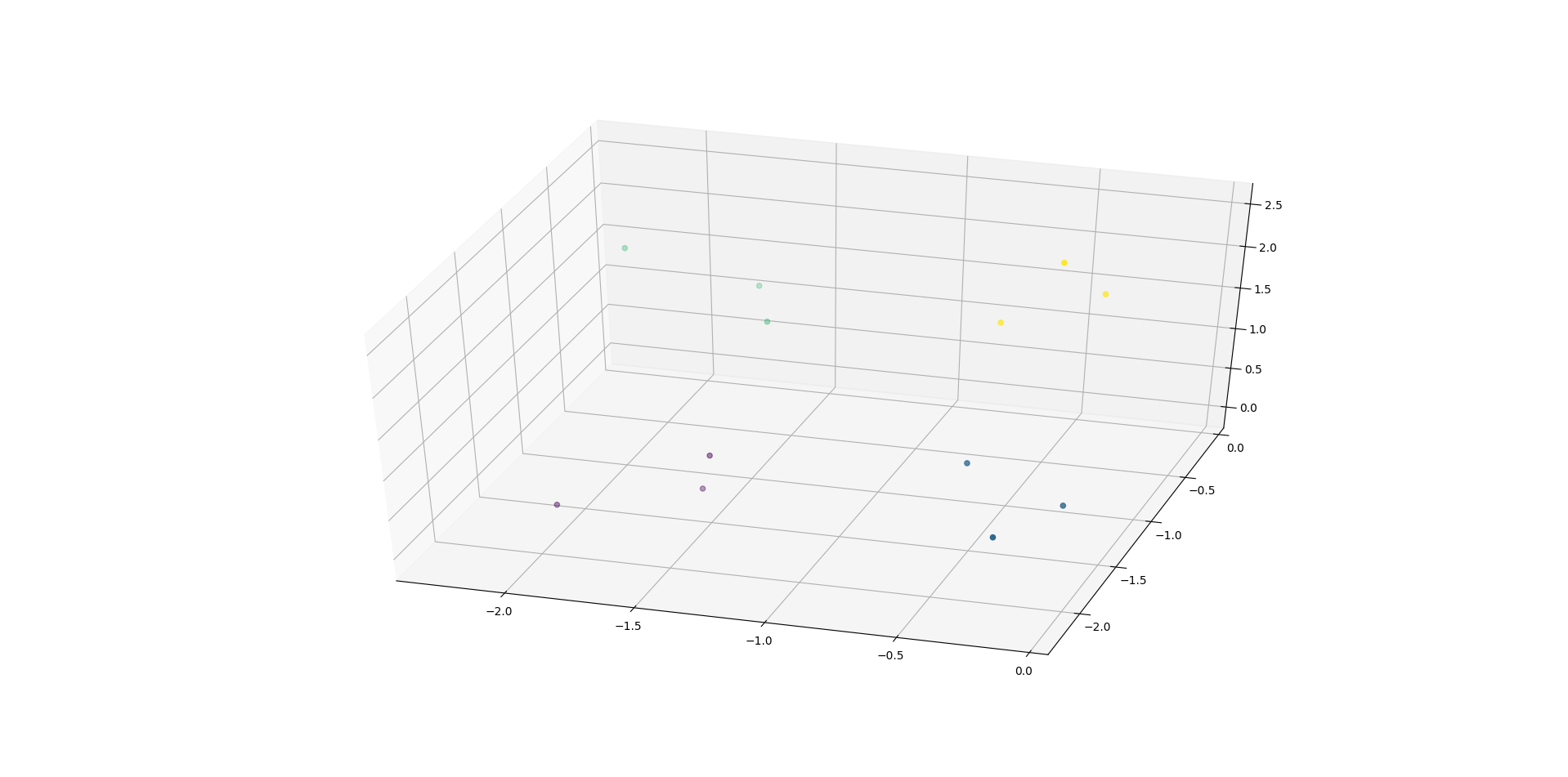}\\
  \includegraphics[width=0.48\textwidth]{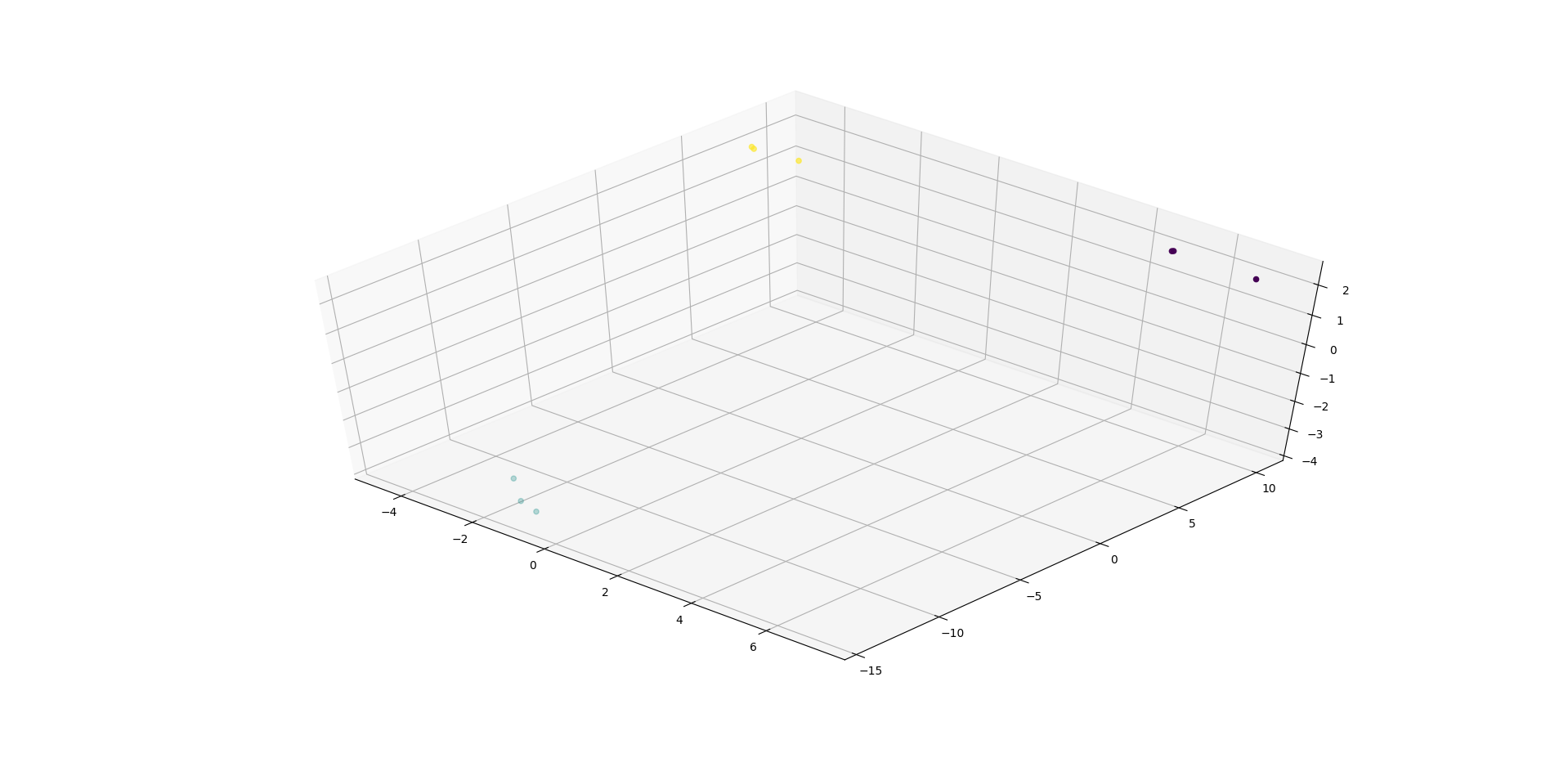}
   \includegraphics[width=0.48\textwidth]{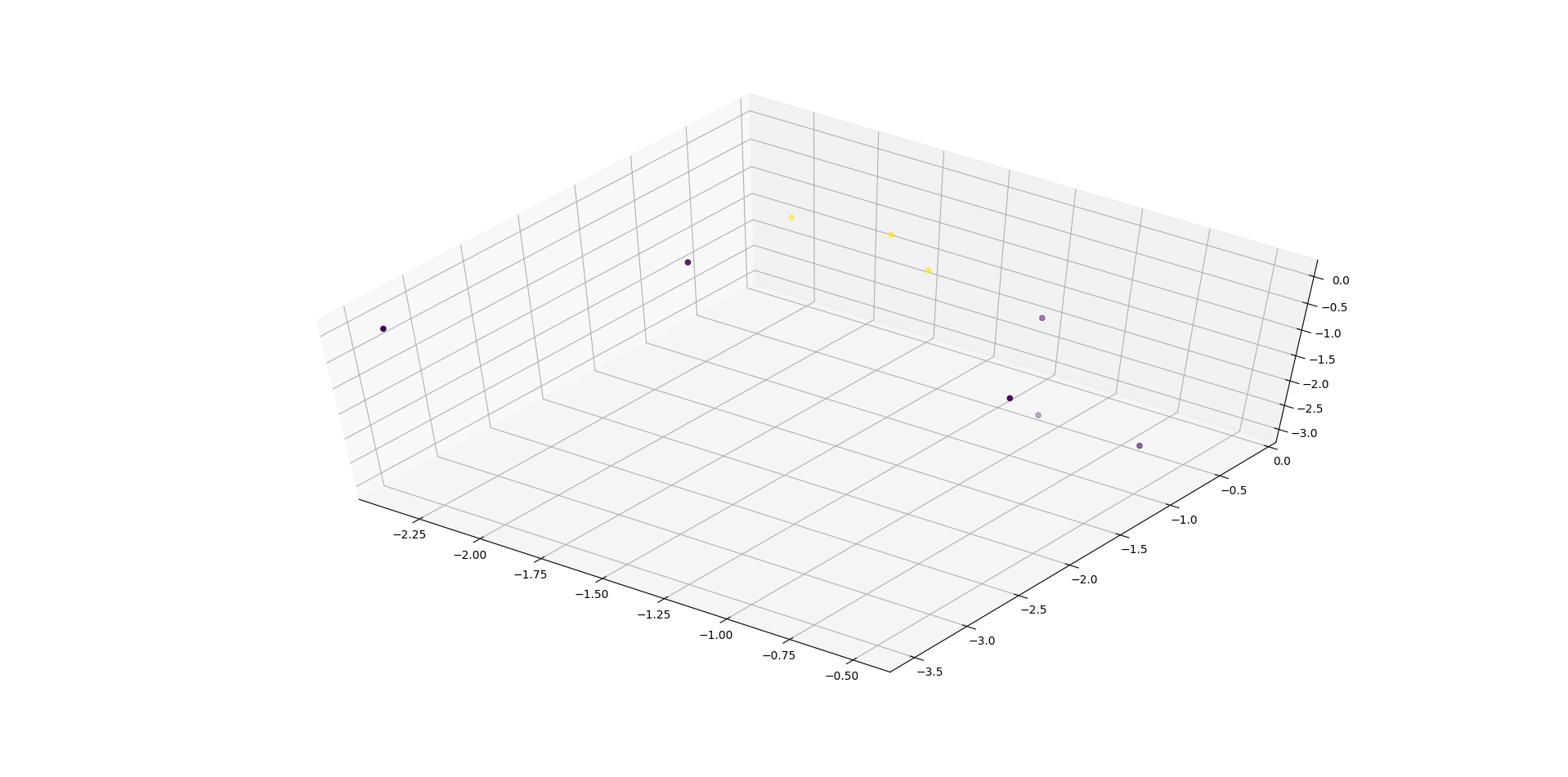}\\
     \includegraphics[width=0.48\textwidth]{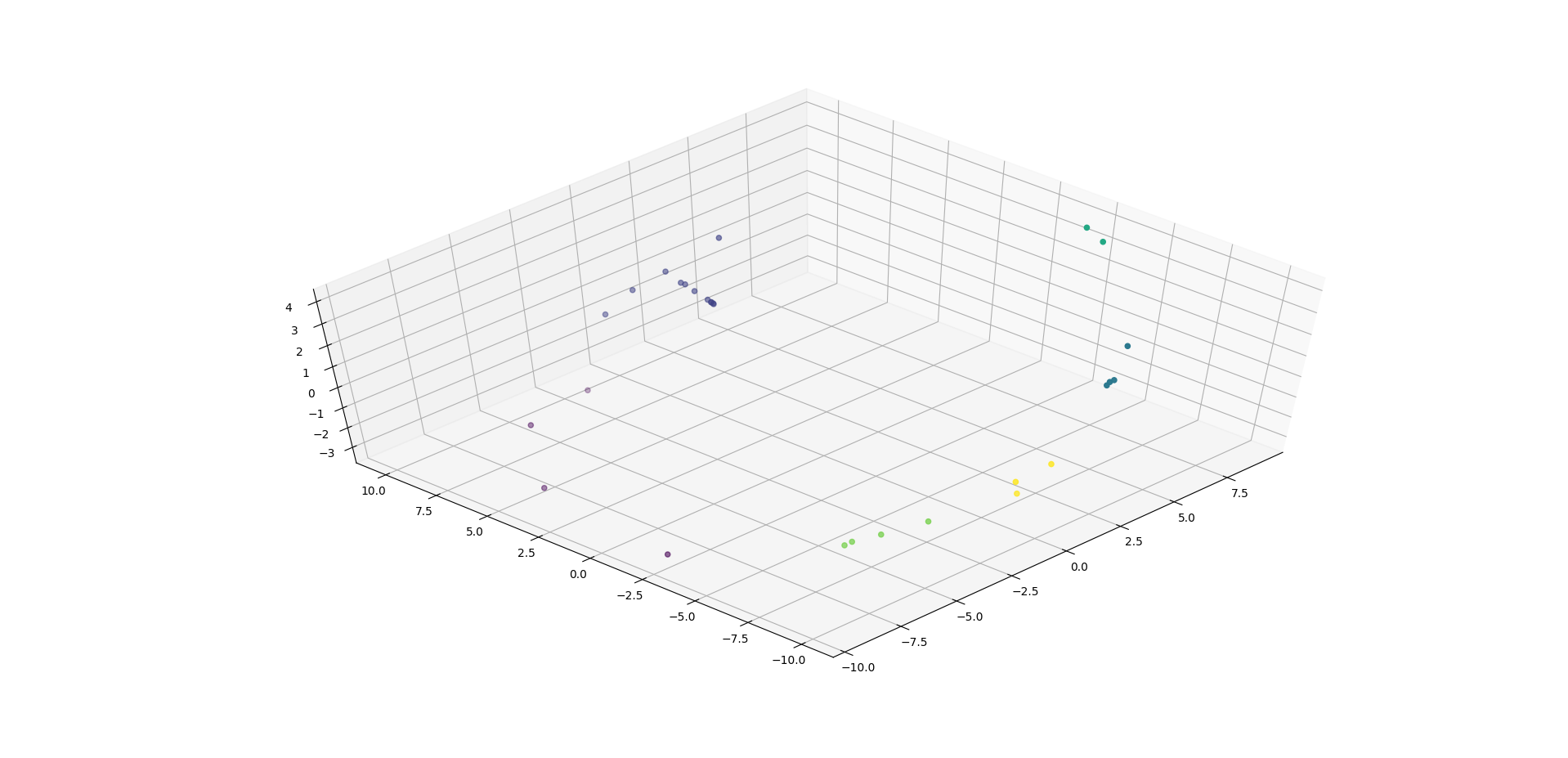}
   \includegraphics[width=0.48\textwidth]{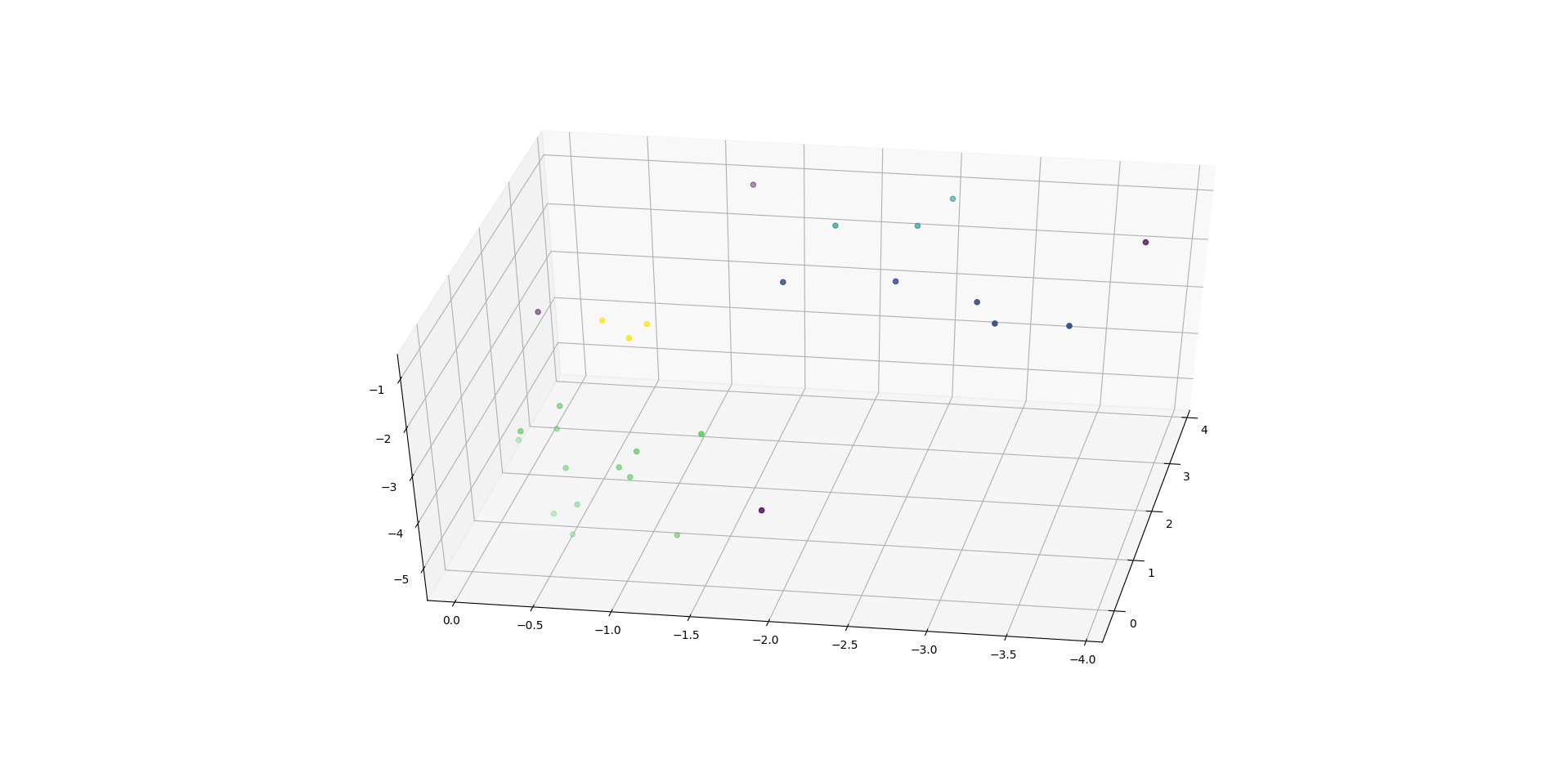}\\
 \caption{
Examples of maps learned for SyncMap (left column) and Word2vec (right column). Rows from top to bottom show tests on respectively fixed chunks, mixed chunks and sequence2.
}
 \label{map}
\end{figure}

Figure~\ref{map} shows the difference between maps learned for both SyncMap and Word2vec. In general, SyncMap's learned map are less widely spread and therefore clusters, if present, are clearer.
Clustering in this space is also made easier.

\section{Appendix C - Time Analysis}

Table~\ref{computation_time} shows the time required to compute each of the methods.
SyncMap is the second fastest after PARSER. 
It is impressive that SyncMap can be faster than word2vec even when word2vec is probably the algorithm  which has the highest investment on efficiency improvement from the algorithms tested. 
Both SyncMap and MRIL should improve considerably if parallelization, GPGPU programming and other techniques are employed. For example, all neurons in SyncMap can be updated at the same time.

\begin{table}
  \caption{Computation time comparison over SyncMap, PARSER, Word2vec and MRIL.}
  \label{computation_time}
  \centering
  \begin{tabular}{lccccc}
    \toprule
    {\bf Model}     & SyncMap & PARSER &Word2vec &MRIL\\
    \midrule
    Computation Time[s] & 8.60$\pm$0.21 & 1.74$\pm$0.05&13.832$\pm$0.54&25.5$\pm$0.56\\

    \bottomrule
  \end{tabular}
\end{table}

\section{Appendix D - SyncMap's Parameter Variations}

Tables~\ref{syncmap_continual}, \ref{syncmap_fixed_long_mixed} and~\ref{syncmap_overlap_sequence} show the performance of SyncMap on the same experiments but with different settings.
Results suggest that SyncMap is robust to changes in parameters, with mostly smooth changes. 

\begin{table}
  \caption{Mutual information of several SyncMap variations in continual settings. The variables inside brackets have the following meanings. The first value is the adaptation rate. The second value modifies the adaptation rate, it can be either fixed (value: fixed) or multiplied by the number of outputs (value: out). The third value indicates the dimensions of map $w$.}
  \label{syncmap_continual}
  \centering
  \begin{tabular}{lccccc}
    \toprule
        \multirow{2}{*}{\bf Model} &
      \multicolumn{2}{c}{\bf Continual fixed} &
      \multicolumn{2}{c}{\bf Continual mixed}\\ 
      \cline{2-5}
    & {Task 1} & {Task 2} &{Task 1} &{Task 2}\\  
    \midrule
    {\bf Ours}: SyncMap (0.1, fixed, 3d)  & 0.93$\pm$0.13 & 0.89$\pm$0.15 & 0.97$\pm$0.04 & 0.99$\pm$0.03\\
    \midrule
    {\bf Ours}: SyncMap (0.01, fixed, 3d)  & 0.99$\pm$0.01 &  1.0$\pm$0.0 & 0.99$\pm$0.03 & 0.92$\pm$0.08\\
    \midrule
    {\bf Ours}: SyncMap (0.01, out, 3d)  & 0.89$\pm$0.11 & 0.85$\pm$0.15 & 0.99$\pm$0.03 & 0.98$\pm$0.05\\
    \midrule
    {\bf Ours}: SyncMap (0.001, out, 3d)  &  0.98$\pm$0.06 & 0.96$\pm$0.18 & 0.98$\pm$0.05 & 0.92$\pm$0.07\\
    \midrule
    {\bf Ours}: SyncMap (0.1, fixed, 2d)  & 0.70$\pm$0.08 & 0.69$\pm$0.06 & 0.97$\pm$0.07 & 0.98$\pm$0.04\\
    \midrule
    {\bf Ours}: SyncMap (0.01, fixed, 2d)  & 0.91$\pm$0.13 & 0.80$\pm$0.34 & 0.96$\pm$0.05 & 0.93$\pm$0.06\\
    \midrule
    {\bf Ours}: SyncMap (0.01, out, 2d)  & 0.81$\pm$0.13 & 0.80$\pm$0.13 & 0.98$\pm$0.04 & 0.99$\pm$0.03\\
    \midrule
    {\bf Ours}: SyncMap (0.001, out, 2d)  & 0.86$\pm$0.17 & 0.70$\pm$0.37 & 0.97$\pm$0.07 & 0.93$\pm$0.08\\
    \bottomrule
  \end{tabular}
\end{table}

\begin{table}
  \caption{Mutual information of several SyncMap variations in Fixed chunks, Long chunks and Mixed structures settings. The variables inside brackets have the following meanings. The first value is the adaptation rate. The second value modifies the adaptation rate, it can be either fixed (value: fixed) or multiplied by the number of outputs (value: out). The third value indicates the dimensions of map $w$.}
  \label{syncmap_fixed_long_mixed}
  \centering
  \begin{tabular}{lccccc}
    \toprule
    {\bf Model}     & {\bf Fixed Chunks}       & {\bf Long Chunks}  &{\bf Mixed Structures}\\
    \midrule
    {\bf Ours}: SyncMap (0.1, fixed, 3d)  & 0.97$\pm$0.09 & 0.97$\pm$0.06 & 0.95$\pm$0.06\\
    \midrule
    {\bf Ours}: SyncMap (0.01, fixed, 3d)  & 0.99$\pm$0.06 &  0.84$\pm$0.13 & 0.98$\pm$0.04\\
    \midrule
    {\bf Ours}: SyncMap (0.01, out, 3d)  & 0.90$\pm$0.13 & 0.96$\pm$0.05 & 0.98$\pm$0.05\\
    \midrule
    {\bf Ours}: SyncMap (0.001, out, 3d)  &  0.96$\pm$0.10 & 0.90$\pm$0.12 & 0.98$\pm$0.05\\
    \midrule
    {\bf Ours}: SyncMap (0.1, fixed, 2d)  & 0.71$\pm$0.10 & 0.97$\pm$0.06 & 0.97$\pm$0.04\\
    \midrule
    {\bf Ours}: SyncMap (0.01, fixed, 2d)  & 0.88$\pm$0.17 & 0.86$\pm$0.12 & 0.97$\pm$0.05\\
    \midrule
    {\bf Ours}: SyncMap (0.01, out, 2d)  & 0.81$\pm$0.14 & 0.97$\pm$0.05 & 0.99$\pm$0.03\\
    \midrule
    {\bf Ours}: SyncMap (0.001, out, 2d)  & 0.88$\pm$0.17 & 0.82$\pm$0.15 & 0.97$\pm$0.06\\
    \bottomrule
  \end{tabular}
\end{table}

\begin{table}
  \caption{Mutual information of several SyncMap variations in Overlapped chunks and Real world scenarios. The variables inside brackets have the following meanings. The first value is the adaptation rate. The second value modifies the adaptation rate, it can be either fixed (value: fixed) or multiplied by the number of outputs (value: out). The third value indicates the dimensions of map $w$.}
  \label{syncmap_overlap_sequence}
  \centering
  \begin{tabular}{lccccc}
    \toprule
    {\bf Model}     & {\bf Overlap1}       & {\bf Overlap2}  &{\bf Sequence1} &{\bf Sequence2}\\
    \midrule
    {\bf Ours}: SyncMap (0.1, fixed, 3d)  & 0.70$\pm$0.19 & 0.64$\pm$0.10 & 0.39$\pm$0.16 & 0.61$\pm$0.06 \\
    \midrule
    {\bf Ours}: SyncMap (0.01, fixed, 3d)  & 0.70$\pm$0.16 &  0.71$\pm$0.10 & 0.34$\pm$0.1 & 0.51$\pm$0.07\\
    \midrule
    {\bf Ours}: SyncMap (0.01, out, 3d)  & 0.80$\pm$0.15 & 0.61$\pm$0.05 & 0.33$\pm$0.11 & 0.56$\pm$0.07\\
    \midrule
    {\bf Ours}: SyncMap (0.001, out, 3d)  &  0.70$\pm$0.14 & 0.74$\pm$0.14 & 0.37$\pm$0.11  & 0.56$\pm$0.07\\
    \midrule
    {\bf Ours}: SyncMap (0.1, fixed, 2d)  & 0.69$\pm$0.22 & 0.62$\pm$0.08 & 0.38$\pm$0.12  & 0.63$\pm$0.05\\
    \midrule
    {\bf Ours}: SyncMap (0.01, fixed, 2d)  & 0.55$\pm$0.26 & 0.67$\pm$0.09 & 0.32$\pm$0.12  & 0.45$\pm$0.07\\
    \midrule
    {\bf Ours}: SyncMap (0.01, out, 2d)  & 0.80$\pm$0.18 & 0.64$\pm$0.09 & 0.34$\pm$0.12  & 0.57$\pm$0.06\\
    \midrule
    {\bf Ours}: SyncMap (0.001, out, 2d)  & 0.55$\pm$0.27 & 0.68$\pm$0.11 & 0.38$\pm$0.12  & 0.55$\pm$0.09\\
    \bottomrule
  \end{tabular}
\end{table}



\bibliographystyle{plain}
\bibliography{general_chunking}